# A Multi-Timestep LSTM Ensemble regressor for Enhanced Short-Term Runoff Prediction

Hamid Saadatfar
Associate Professor, Department of Computer Engineering, Faculty of Electrical and Computer Engineering, University of Birjand, Birjand, Iran.
Email: saadatfar@birjand.ac.ir

AmirHossein Eshghi (corresponding author)
Non-degree Student, Department of Computer Science, Clemson University, Clemson, SC, USA.

M.Sc. Student, Department of Computer Engineering, Faculty of Electrical and Computer Engineering, University of Birjand, Birjand, Iran.

Email: aeshghi@Clemson.edu, amir.eshghi@birjand.ac.ir

Behnaz Behdani
BENG student, Department of Computer Engineering, Faculty of Electrical and Computer Engineering, University of Birjand, Birjand, Iran.
Email: behnazbehdani@outlook.com

**Abstract**

Accurately forecasting river runoff is key to managing water resources, controlling floods, and planning agriculture. This study looks at the Ajichay River in northwest Iran, a major tributary of Lake Urmia that has seen increasing water-related stress in recent years. We introduce a daily runoff prediction model based on Long Short-Term Memory (LSTM) networks. The model combines five LSTM units, each trained on different time intervals (from 2 to 6 days), to better capture the river's changing flow patterns. To improve performance, we fine-tuned each model using Particle Swarm Optimization (PSO), a population-based algorithm. We tested the model on unseen data from 2017–2018, using $R^2$, RMSE, and MSE as evaluation metrics. Results showed strong accuracy, with $R^2$ scores ranging from 74.95% to 91.42%. We also used different methods to identify the most important features, offering deeper insights into what drives runoff changes.

**Keywords:** Runoff prediction, Water resource, LSTM, Analysis data.

## 1-Introduction

Optimal water resource management stands as on the primary challenges of our era, playing a decisive role in sustainable development, food security and the diminution of natural hazards[1]. Among these efforts, precise river flow prediction, as a key indicator in water resource planning, has gained an increasing importance[2]. The ability to estimate the approximate amount of runoff is not only beneficial to efficient allocation of water for agricultural, drinking and industrial purposes, but also serves as a vital tool in flood management and the reduction of damages caused by river overflows[3, 4].

Given climate change and the rising fluctuation of precipitation patterns in the past recent years, the need for dynamic and accurate modeling of river flow has been more pressing than ever[5]. In the warmer seasons, this precise forecasting of available water, particularly in arid or semi-arid regions, enables planning for sustainable agriculture and ensure food security[6]. On the other hand, during colder and rainy seasons, accurate runoff forecasting can prevent human and economic disasters caused by river overflows[7].

Today with the advancements that we have in artificial intelligence (AI), machine learning (ML) [8] and remote sensing thechnologies, new methods for hydrological modeling have been introduced[9]. Machine learning techniques have revolutionized hydrological modeling by not only tremendously enhancing the accuracy of river flow prediction [10] but also by enabling the identification of key environmental factors influencing runoff through the analysis of complex datasets. Additionally, deep learning (DL)[11] techniques such as LSTM (Long ShortTerm Memory)[12] and CNN (Convolutional Neural Networks)[13] , can extract hidden patterns in river behaviors through processing spatiotemporal data. An LTSM model can detect long-term dependencies between seasonal precipitation and river discharge, an ability beyond most traditional methods. This capacity is particularly critical in watersheds with highly variable climatic conditions, where numerous factors nonlinearly influence runoff[14].

Among the strongest neural network architectures for processing spatiotemporal data such as runoff prediction, LSTM networks stand out for their ability to learn long-term dependencies and complex temporal patterns, thanks to their unique gating mechanisms[15]. The implication of these mechanisms is often ignored in other ordinary neural models and networks[16]. Especially in runoff forecasting, where parameters such as precipitation, temperature, and soil moisture exhibit delayed effects, LSTM can accurately model the nonlinear, multiscale relationships among these variables. In this study, to optimize prediction accuracy, an advanced hybrid ensemble approach has been employed that is designed based on an intelligent combination of many LSTM architectures with different configurations[17]. This novel framework can model the inherent complexities of hydrological systems with high precision.

This article is structured as follows: Section 2 comprehensively reviews related studies on river flow prediction; Section 3 consists of the detailed hydrological and meteorological datasets used, as well as the geographic and environmental characteristics of the study area; Section 4 thoroughly explains the proposed methodology; and Section 5 presents and analyzes the experimental results and the impact of various features on runoff.

## 2-Related Work

In recent years, the prediction of runoff tempo using artificial intelligence methods has increasingly attracted the attention of hydrology researchers. Numerous studies have demonstrated that combining signal decomposition techniques with deep neural network architectures can effectively overcome the challenges posed by the nonlinear and nonstationary nature of runoff data[18]. For instance, recent research has employed hybrid strategies based on Empirical Mode Decomposition (EMD) and its improved variants such as CEEMDAN and ICEEMDAN, along with Variational Mode Decomposition (VMD), to reduce noise and extract frequency components. Xu et al [19]. for example, reported significant improvements in monthly runoff prediction accuracy by combining CEEMDAN and VMD for pre-processing data and optimizing LSTM parameters using the CABES algorithm. This approach not only manages high-frequency noise more efficiently, but also takes model parameter tuning to a new level through intelligent optimization mechanisms.

In addition to signal decomposition techniques, the use of metaheuristic algorithms for the simultaneous optimization of prediction model parameters and signal decomposition structures has emerged as a novel trend. Wang et al [20]. demonstrated that combining the Whale Optimization Algorithm (WOA) with VMD and GRU models can simultaneously identify optimal decomposition and prediction parameters. Similarly, He et al [21]. introduced an improved version of the Marine Predators Algorithm (MARO) and integrating it with a Support Vector Regression (SVR) model, successfully achieving a prediction error reduction of up to 80.6%. These achievements show that the intelligent integration of optimization methods with deep learning architectures can significantly improve models' capabilities to learn complex hydrological patterns.

The development of hybrid convolutional-neural architectures has also emerged as a promising research direction. Zhang [22], by introducing the EA-TCN framework, which combines Temporal Convolutional Networks (TCNs) with lightweight attention mechanisms and group learning, achieved superior short-term prediction accuracy. Meanwhile, Hu [23], in a study conducted on the Yellow River Basin, demonstrated the effectiveness of CNN-LSTM combinations in simultaneously extracting spatiotemporal features from multi-source gridded data. These approaches have notably improved model performance, particularly in data-scarce environments, by integrating spatial information on precipitation and soil moisture.

Attention mechanisms and error correction strategies have also been employed as key components to enhance model reliability. Recent advancements in post-processing techniques, such as attention mechanisms and error correction strategies, have significantly enhanced the reliability of predictive models. Wang et al [24]. for instance, introduced the WD-AM-LSTM framework, which incorporates interpretable attention layers alongside wavelet-based noise reduction. Their approach not only improved predictive accuracy but also reduced computational costs by up to 92%. In other studies, Feng et al [25]. developed the TELM-CSA model for multi-step point and interval forecasting, enabling the simultaneous estimation of confidence intervals—an important step towards improving the trustworthiness of model outputs. These advancements demonstrate that integrating post-processing strategies, such as error correction and interpretability mechanisms, can bridge the gap between complex deep learning models and the practical demands of water resource management.

Nevertheless, substantial challenges remain, particularly regarding the optimal tuning of signal decomposition parameters, the integration of heterogeneous data sources, and the enhancement of model interpretability. Recent work by Xu et al [26]. employing stochastic walk strategies and elite opposition-based learning in optimization algorithms, has opened new paths for addressing issues like local optimization. Moreover, the design of multi-stage frameworks—such as the DVMD-DBO-LSSVM-EC model, which incorporates error correction as a final stage—underscores the value of multi-phased approaches for improving final prediction accuracy. Collectively, these developments highlight the critical importance of hybrid strategies that combine the strengths of signal decomposition techniques, advanced optimization algorithms, and innovative neural network architectures.

## 3. Study Area and Data

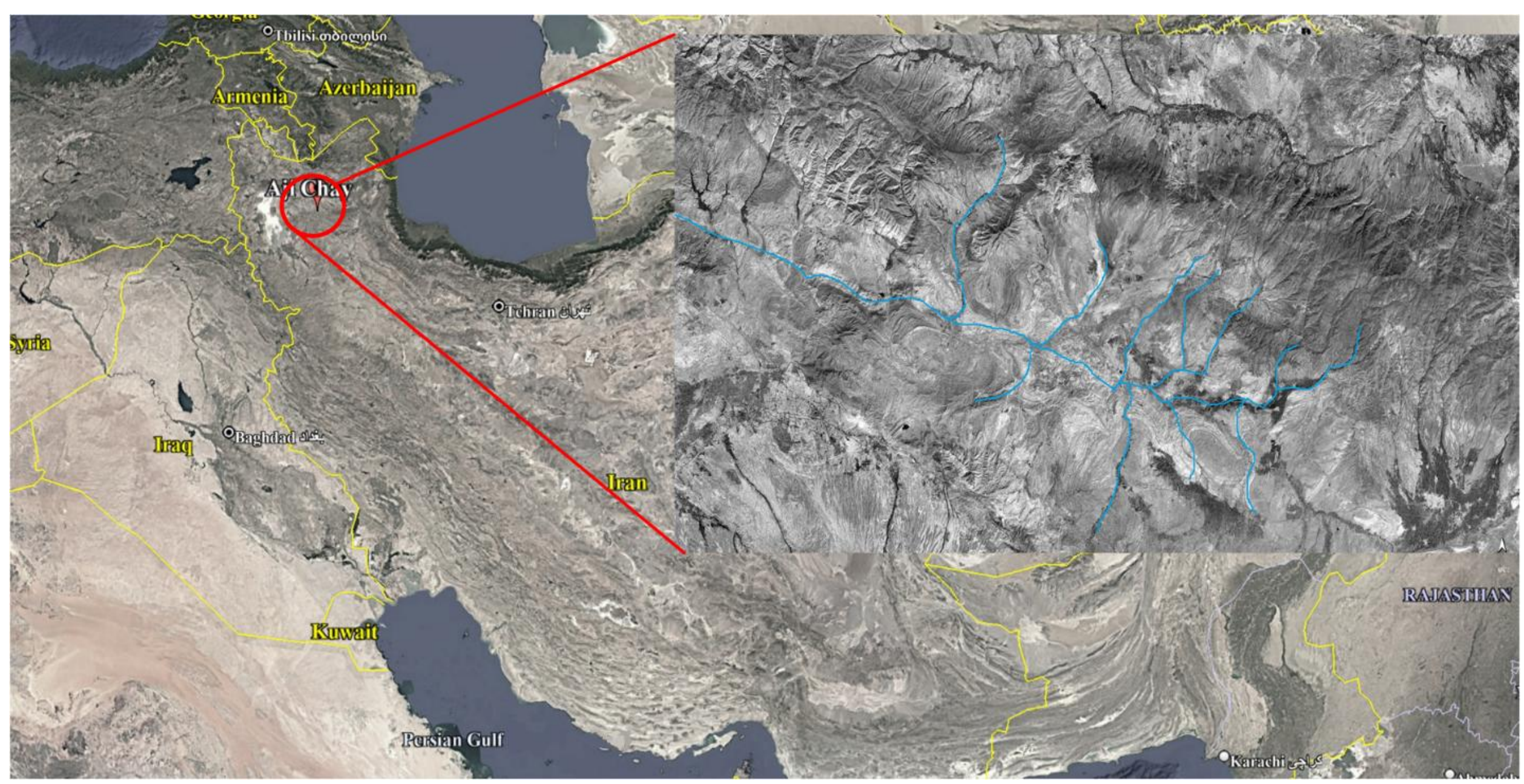

Figure 1. Geographical location of the Ajichai River

The Ajichay River (Figure 1), one of the most important tributaries feeding Lake Urmia, is located between 37° and 39° north latitude and 45° and 46° east longitude[27]. Positioned to the northwest of Lake Urmia (Figure 2), it plays a crucial role in sustaining the water resources of this ecologically sensitive region. Climatically, the Ajichay watershed receives an average annual precipitation of approximately 350 millimeters, with nearly 20% falling during March and April. The area experiences pronounced temperature variability: the mean annual temperature is 7.8°C, while extremes range from a maximum of 31°C in July to a minimum of –15.5°C in February. Given the concentration of precipitation during the spring months and the watershed's steep topography, the risk of seasonal flooding especially in downstream areas is extremely high. With numerous urban and rural communities situated along its banks, the effective management of runoff and the accurate forecasting of river flow are matters of both economic and security concern.

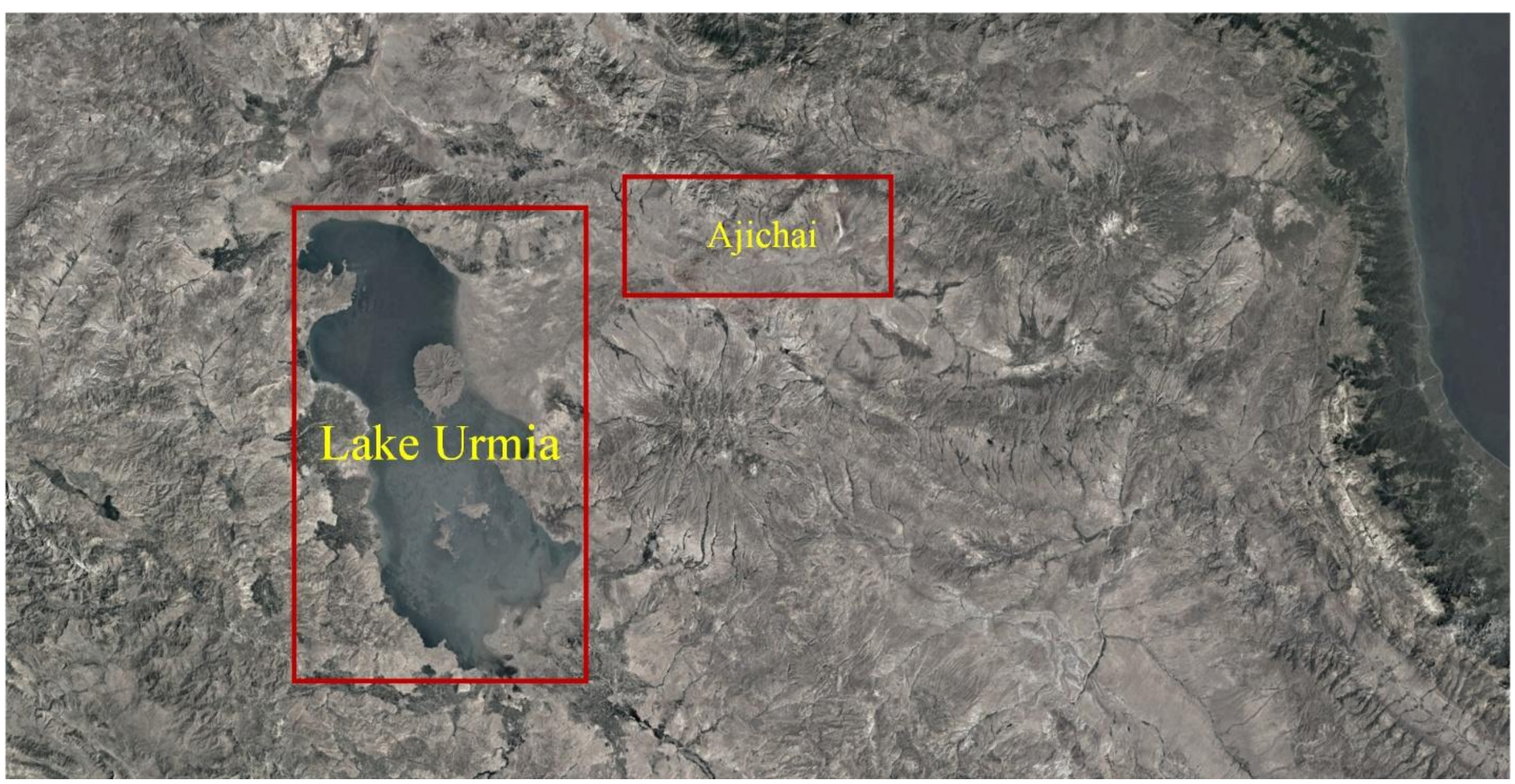

Figure 2. Relative location of Ajichai to Lake Urmia

This study utilizes a dataset comprising 1,604 samples collected between 2010 and 2018 through meteorological stations and satellite image analysis. Given the focus on forecasting river flow during the first half of the year (January to June) a period marked by heightened precipitation and flood risk the dataset was confined to this seasonal window. Data were drawn from four stations strategically selected along the Ajichay River according to elevation gradients, from upstream to downstream. Key variables include hydrological and meteorological parameters influencing river discharge, with descriptive statistics (minimum, maximum, and mean) detailed in Table 1. This sampling strategy allows for a comprehensive spatiotemporal assessment of river behavior under critical precipitation conditions.

Table 1. Feature information in the dataset

| **Feature Name** | **Average** | **Maximum** | **Minimum** |
| --- | --- | --- | --- |
| Avg Temp Z1 (C) | 7.88 | 29.6 | -15.8 |
| Avg Temp Z2 (C) | 5.93 | 27.65 | -17.75 |
| Avg Temp Z3 (C) | 2.68 | 24.40 | -21.00 |
| Avg Temp Z4 (C) | -1.06 | 20.65 | -24.75 |
| Snow cover Z1 (%) | 0.06 | 0.97 | 0.0 |
| Snow cover Z2 (%) | 8.7 | 99.48 | 0.0 |
| Snow cover Z3 (%) | 10.27 | 99.54 | 0.0 |
| Snow cover Z4 (%) | 11.34 | 99.67 | 0.0 |
| Precipitation Z1 (mm) | 0.14 | 4.1 | 0.0 |
| Precipitation Z2 (mm) | 0.16 | 4.67 | 0.0 |
| Precipitation Z3 (mm) | 0.18 | 5.3 | 0.0 |
| Precipitation Z4 (mm) | 0.20 | 6.02 | 0.0 |

| Month | - | 6 | 1 |
|---|---|---|---|
| Year | - | 2018 | 2010 |
| Previous day Runoff ( $m^3/s$ ) | 6.1 | 86.3 | 0.0 |

To examine inter-annual variations in hydroclimatic parameters, annual averages of snow cover (percentage), precipitation (mm), and temperature (°C) were calculated for each of the four study regions and illustrated in Figure 3. These graphs provide an integrated view of temporal trends, spatial differences among regions, and correlations between key variables over the 2010–2018 period. Notably, the analyses reveal significant differences in snow cover and precipitation patterns between upstream and downstream zones, insights that are crucial for improving the accuracy of river flow forecasts during peak discharge seasons.

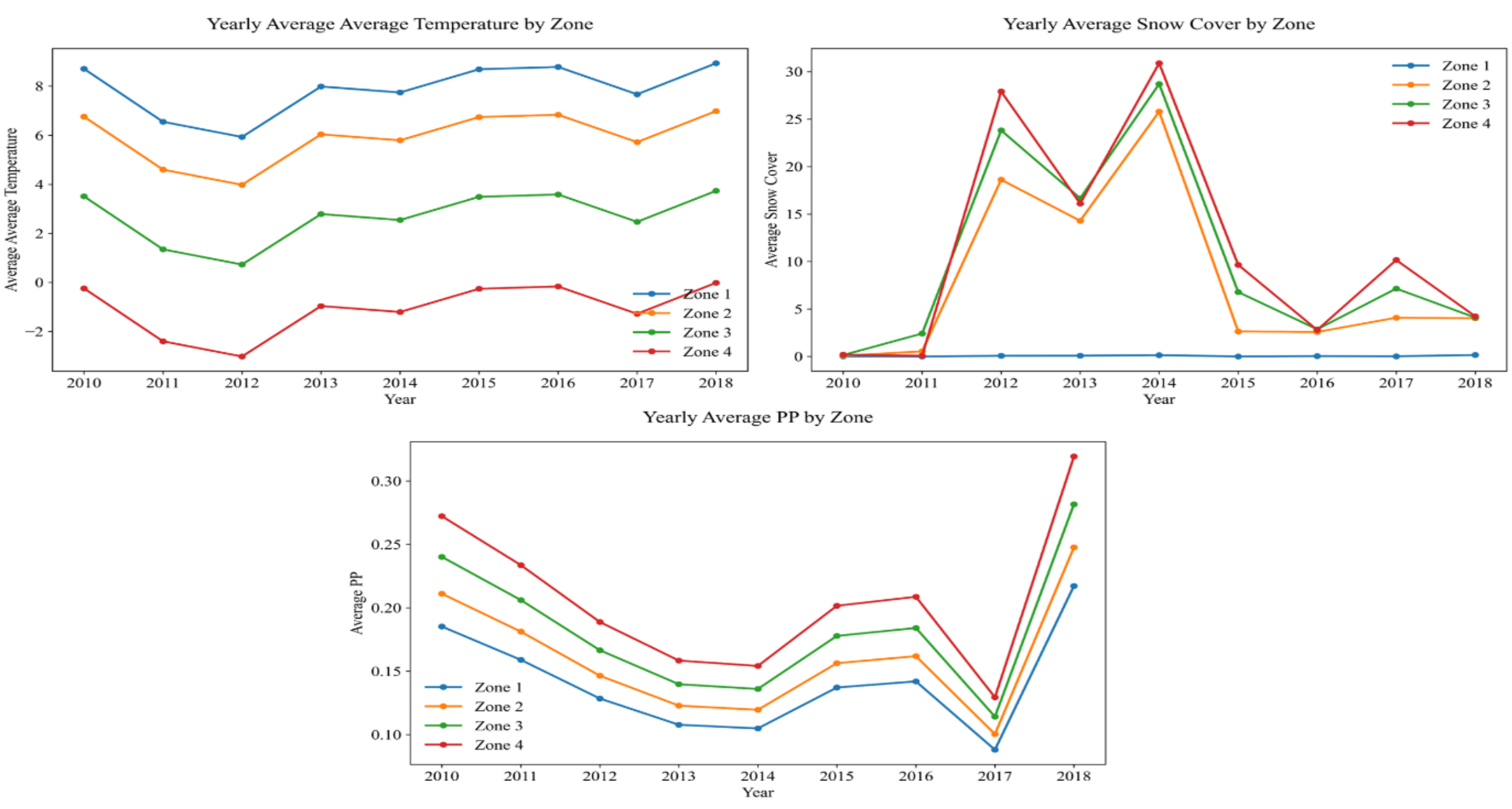

Figure3. Display each parameter by region and year

## 4- Methodology

In this section, we present a detailed explanation of the methodology employed in the study, along with the reason for the selection of specific techniques. At the outset, it must be noted that the prediction task undertaken in this research was approached through the application of a neural network architecture. Deep neural networks, as among the most sophisticated frameworks in machine learning, have demonstrated exceptional capabilities in modeling complex, nonlinear problems[28]. Among these architectures, specialized networks for temporal data occupy a distinctive position, as they are specifically designed to capture complex temporal patterns. Unlike traditional methods, which typically rely on instantaneous, single-step data inputs, temporal neural networks leverage long-term memory mechanisms to identify and model dependencies across multiple preceding time steps. Among these, (LSTM) networks stand out for their effectiveness in temporal prediction tasks [29]. Owing to their distinctive gated architecture—comprising forget, input, and output gates LSTM networks are capable of selectively retaining pertinent information over time, thereby mitigating the vanishing gradient problem that commonly impairs conventional recurrent neural networks (RNNs)[30].

As mentioned, in this study we employed a novel and advanced LSTM-based architecture featuring multiple temporal steps, aimed at overcoming the limitations associated with conventional single-step models (Figure 4). While a standard LSTM with a fixed time step can capture only a narrow range of temporal dependencies, our proposed system integrates multiple LSTM networks, each configured with different temporal windows ranging from two to six days. This flexible design allows the model to concurrently capture both short-term influences, such as recent precipitation, and longer-term processes, such as gradual snowmelt, which collectively govern hydrological behavior. Each individual network is specialized to focus on a specific temporal scale, and their outputs are subsequently combined through intelligent fusion layers. This comprehensive and layered architecture not only significantly enhances prediction accuracy but also substantially improves the model's adaptability to dynamic hydrological conditions. As a result, the system can accurately forecast both the immediate effects of heavy rainfall and the cumulative impacts of environmental factors.

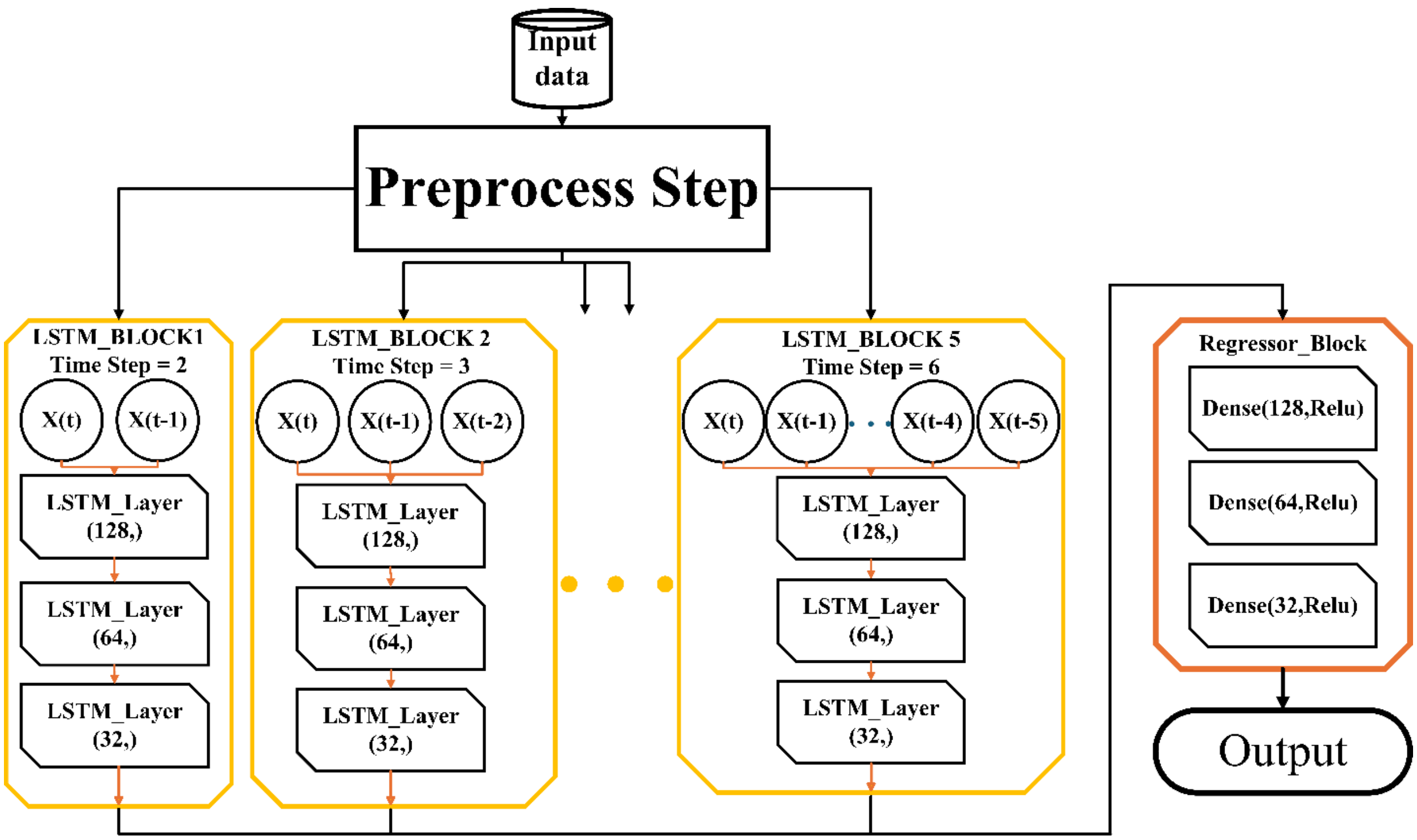

Figure 4 - The main ensemble regressor of the proposed method

## 4-1 Preprocessing

One of the most critical stages in developing an effective machine learning model for accurate prediction is data preprocessing. At the beginning of this study, to preserve the cyclical nature of the “month” variable (ranging from 1 to 12), trigonometric transformations were applied. Two new features, Month_sin and Month_cos, were created using Equation (1), where m denotes the month number.

$$Month_\sin = \text{Sin}\left(\frac{2\pi(m)}{12}\right), \quad Month_\cos = \text{Cos}\left(\frac{2\pi(m)}{12}\right) \tag{1}$$

This transformation has two important aspects. First, it prevents the artificial discontinuity between consecutive months such as December (month 12) and January (month 1). Second, it preserves the nonlinear relationships between months in a manner interpretable by neural networks. Traditional methods, which encode months as discrete values, inadvertently introduce artificial distance between temporally adjacent months.

Following this, all features and the target variable were normalized to the range [0, 1] using the MinMaxScaler algorithm [31] (Equation 2):

$$X_{scaled} = \frac{x - x_{\min}}{x_{\max} - x_{\min}} \tag{2}$$

The rationale behind this scaling is to prevent features with larger numeric ranges from dominating the learning process and to improve compatibility with the activation functions used in neural networks.

Additionally, two new features were engineered and added to the dataset. The first feature, “Runoff Avg 3”, represents the three-day moving average of runoff, providing a smoothed trend of recent runoff behavior. The second feature, “Runoff Std 3”, measures the standard deviation of runoff over the past three days, capturing short-term fluctuations. These two engineered features enhance the model’s capacity to recognize short-term patterns and improve prediction accuracy.

## 5. Results and Experiments

As explained in Section 4, the core idea of this study is to develop an ensemble architecture based on LSTM models that allows temporal tuning of models across different time windows. This structure enables leveraging predictions over various time intervals, ultimately improving the overall accuracy of river flow forecasting. In this section, we present a comprehensive evaluation of the models used, the methods employed for model selection, and the analysis of the influence of different features on model performance.

Firstly, it is important to describe the data partitioning process. For all experiments, data from 2010 to 2016 were used for model training, while data from 2017 and 2018 served as the test set. The reason for this division lies in the nature of temporal forecasting, where evaluating model performance on a truly unseen period provides a more realistic measure of predictive capabilities. Consequently, the years 2017 and 2018 were selected as the test period to assess the model's real-world performance in runoff forecasting.

**5.1. Evaluation Metrics**

To assess prediction accuracy and quantify errors in the regression models, several metrics were employed. In this study, three primary evaluation indexes were utilized: Mean Squared Error (MSE), Root Mean Squared Error (RMSE), and the Coefficient of Determination ($R^2$), each offering distinct advantages for performance analysis.

Mean Squared Error (MSE) [32] (Equation 3) calculates the average of the square differences between the actual and predicted values. Due to squaring, this metric is highly sensitive to large errors and is effective for controlling overall model precision.

$$MSE = \frac{1}{N}\sum_{i=1}^{N}(y_i - \hat{y}_i)^2 \quad (3)$$

Root Mean Squared Error (RMSE) [32] (Equation 4) is derived by taking the square root of MSE. Its key advantage lies in its interpretability, as its units match those of the target variable, thereby facilitating performance comparisons.

$$RMSE = \sqrt{MSE} \quad (4)$$

Coefficient of Determination ($R^2$) [32] (Equation 5) measures the proportion of the variance in the dependent variable that is predictable from the independent variables. This standard metric, ranging from 0 to 1, reflects how well the model captures actual data variability.

$$R^2 = 1 - \frac{\sum_{i=1}^{N}(y_i - \hat{y}_i)^2}{\sum_{i=1}^{N}(y_i - \bar{y}_i)^2} \quad (5)$$

In these equations, y represents the actual value, $\hat{y}$ denotes the predicted value, $\bar{y}$ is the mean of the observed values, and N is the total number of samples in the dataset.

## 5.2. Prediction Model Selection

As previously noted, runoff forecasting, being inherently dependent on historical data, demands the use of powerful temporal models. In this study, LSTM architecture was selected as it stands among the most effective deep learning methods for processing temporal data. Since the choice of an optimal time step plays a crucial role in model performance, various time steps ranging from 2 to 20 days were evaluated. The results of these evaluations are presented in Figure 5-A, which illustrates the impact of different time steps on model prediction accuracy.

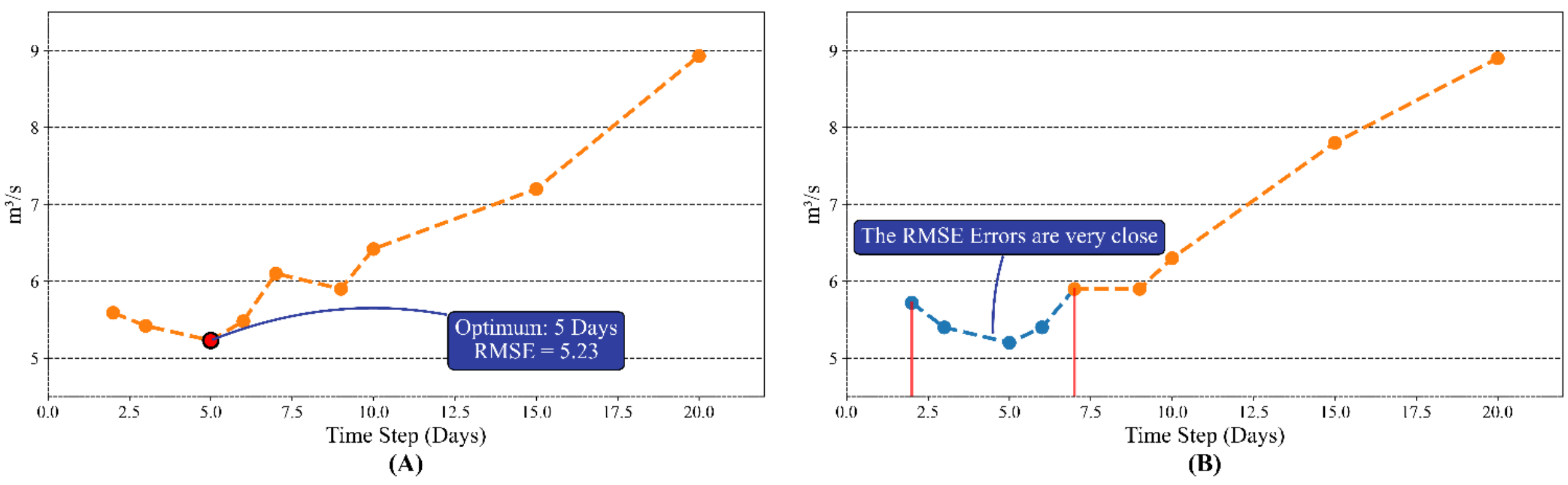

Figure 5 - Results of different experiments to calculate the optimal number of time steps for LSTM

The optimization results indicated that the optimal time step was 5 days, suggesting that information from the preceding five days had the most impact on runoff prediction. However, it was noteworthy that time windows ranging from 2 to 7 days produced very similar errors, implying that all these recent days significantly influence predictions. Figure 5-B indicates that within this range, each day's data plays a substantial role in the final prediction. On the other hand, runoff is influenced by various climatic factors such as snowfall and temperature, each potentially impacting it in distinct ways over different time horizons. In such a scenario, a single LSTM model with a fixed temporal structure may not adequately capture the diverse and complex effects of past days on runoff behavior. To address this limitation and enhance the model's predictive performance, an ensemble architecture comprising multiple LSTM models was developed. This ensemble approach enables a more detailed examination of the contribution of each day within the considered time window, ultimately leading to improved forecast accuracy.

To determine the optimal number of LSTM blocks in the ensemble, a series of experiments were conducted. In each experiment, the number of blocks was progressively increased until the best configuration was identified. Importantly, during each iteration, the Particle Swarm Optimization (PSO) algorithm was employed to find the optimal time step for each individual LSTM block. By intelligently searching through the parameter space, the PSO algorithm proposed the most effective time step for each model separately (Figure 6). The time step values optimized through this process were subsequently used in the final model configuration, ensuring that each block operated with parameters best suited to enhance the overall prediction performance.

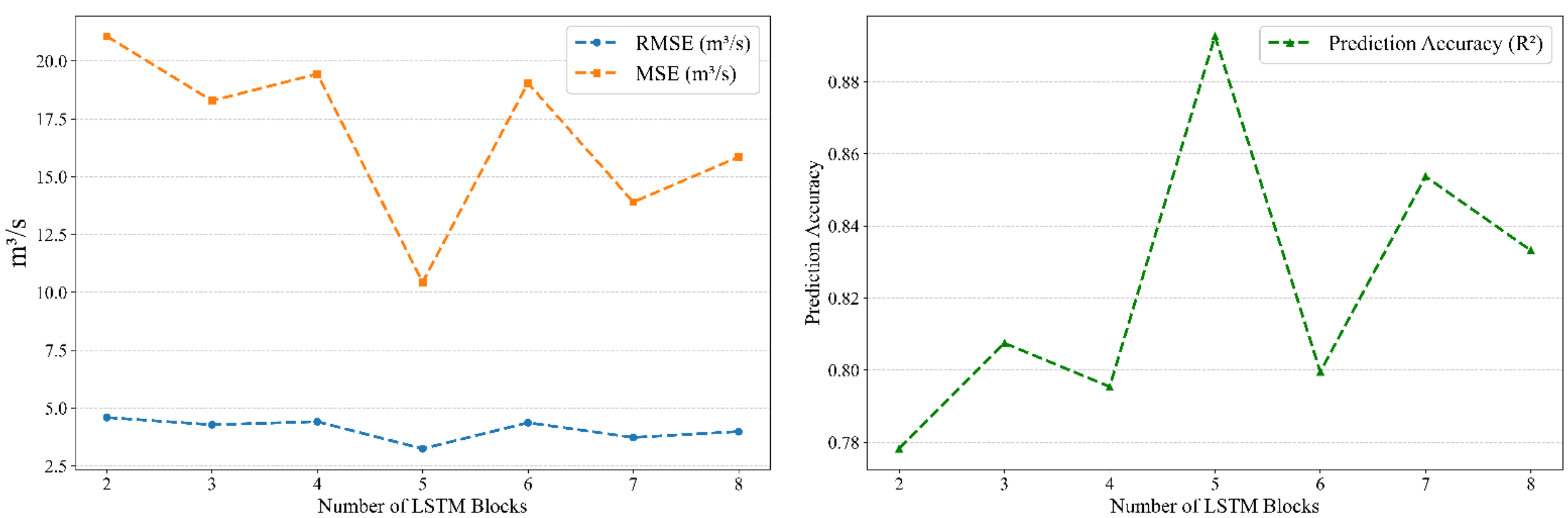

Figure 6 – Results obtained with the PSO algorithm to find the optimal number of LSTM blocks in the best time step with different evaluation criteria

The results demonstrated that an ensemble structure comprising five LSTM blocks significantly outperformed other configurations, providing a marked improvement in predictive accuracy. Consequently, five blocks were selected for the final architecture. In Figure 7, the comparison between the actual and predicted runoff values over the test period (2017 and 2018) is presented, illustrating the model's high accuracy in the tests' time period.

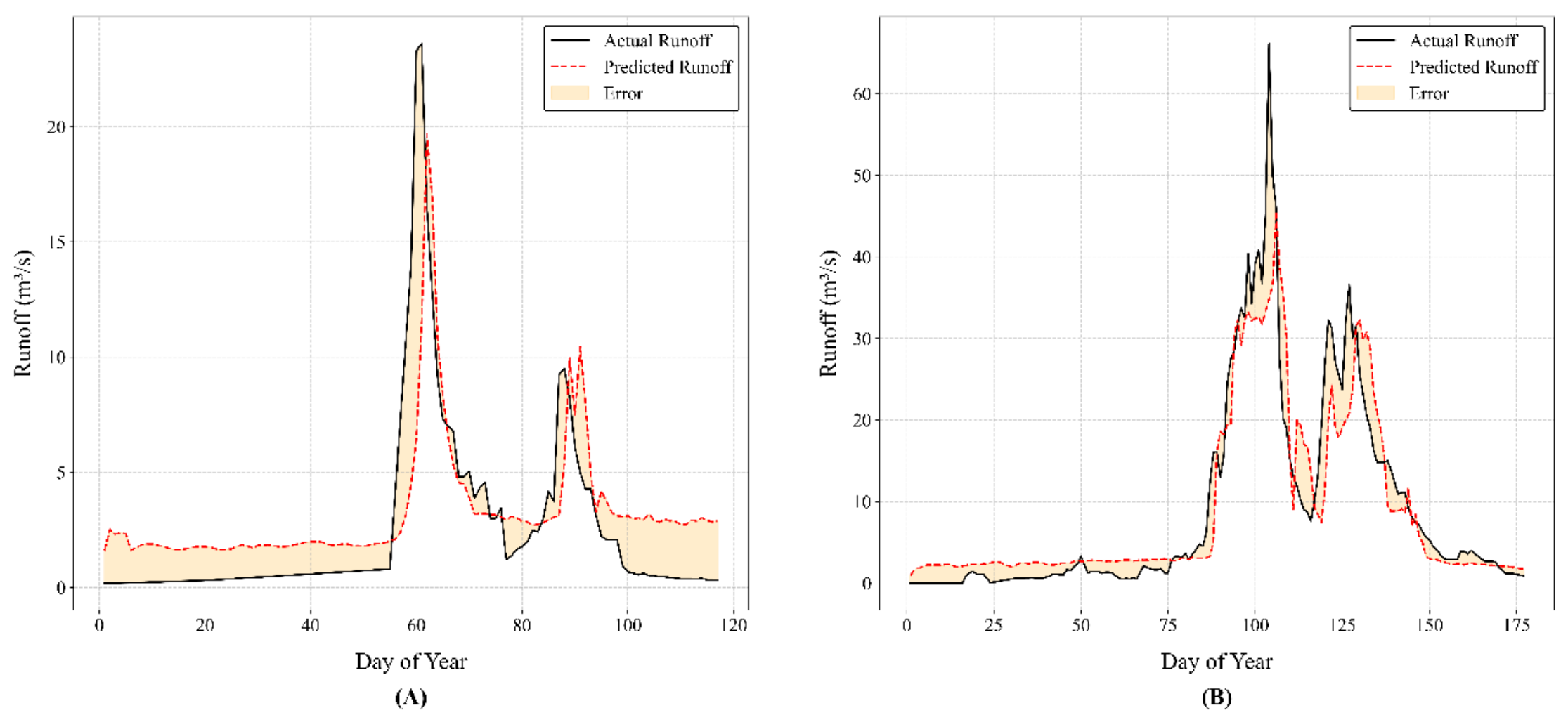

Figure 7 – Comparison between predicted and actual runoff values

(A) Year 2017 - (B) Year 2018

To evaluate the extent of overfitting in long-term predictions, it was necessary to also examine the model's performance over short-term intervals. For this purpose, four months were randomly selected from the years 2017 and 2018. The data corresponding to each selected month was fed separately into the model to generate forecasts. This experiment allowed us to assess the model's ability to perform under different conditions and over short timeframes. The results of the comparisons between actual and predicted values for these four months are illustrated in Figure 8.

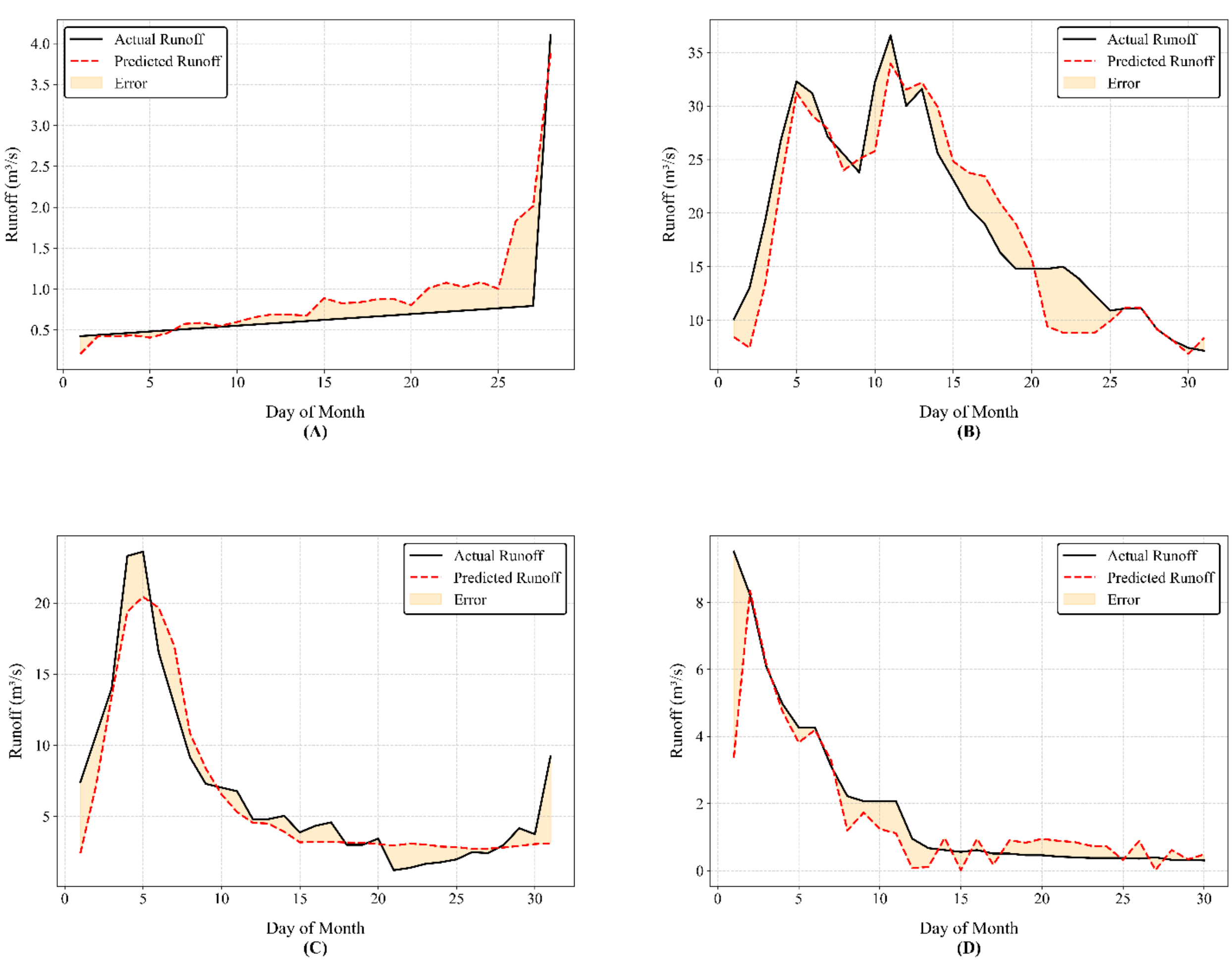

Figure 8. Comparison of observed versus predicted daily runoff values for four randomly selected months: (A, B) from 2017 and (C, D) from 2018, demonstrating the model's short-term forecasting accuracy under varying seasonal conditions.

The months were selected entirely at random. For a more detailed explanation of Figure 8, subfigures (A) and (B) correspond to the data from 2017, while subfigures (C) and (D) pertain to the data from 2018. The figure provides a side-by-side comparison of the observed and predicted runoff values across short-term periods.

Additionally, Table 2 summarizes the results of both annual and monthly predictions. The table reports the evaluation metrics, including $R^2$, RMSE, and MSE, highlighting the final performance of the proposed ensemble model. The obtained values demonstrate the model's high accuracy in forecasting river runoff across both short-term and long-term scales.

Table 2. Prediction results of the ensemble model for short-term and long-term horizons

| Detail | RMSE ($m^3/s$) | MSE ($m^3/s$) | $R^2$ (%) |
|---|---|---|---|
| Year (A) | 2.42 | 5.88 | 81.39 |
| Year (B) | 3.64 | 13.26 | 91.42 |
| Month (A) | 0.35 | 0.12 | 89.48 |
| Month (B) | 3.34 | 11.17 | 84.84 |
| Month (C) | 2.21 | 4.89 | 85.41 |
| Month (D) | 1.21 | 1.47 | 74.95 |

Finally, Figure 9 displays the error histograms separately for the training and test sets from 2010 to 2018, while Figure 10 presents the overall error distribution across all data.

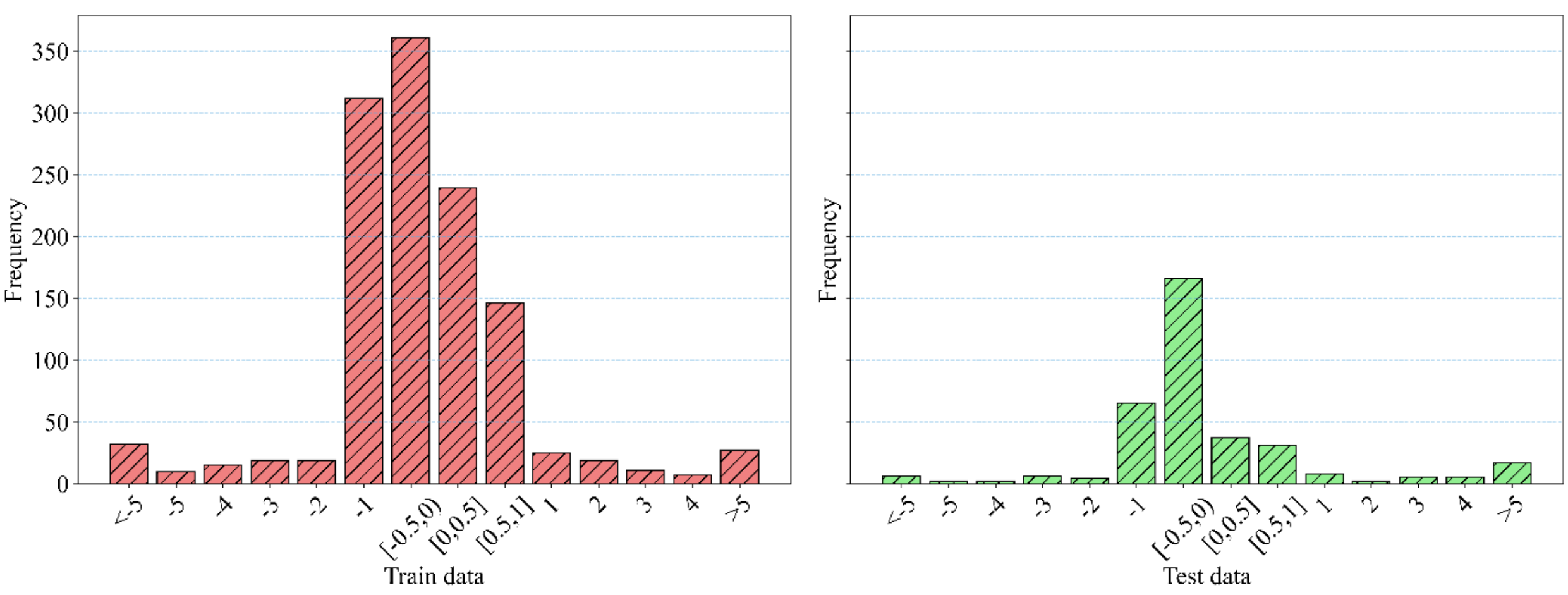

Figure 9. Error distribution histograms for (A) training (2010–2016) and (B) test sets (2017–2018), illustrating the frequency of prediction errors across different runoff magnitudes.

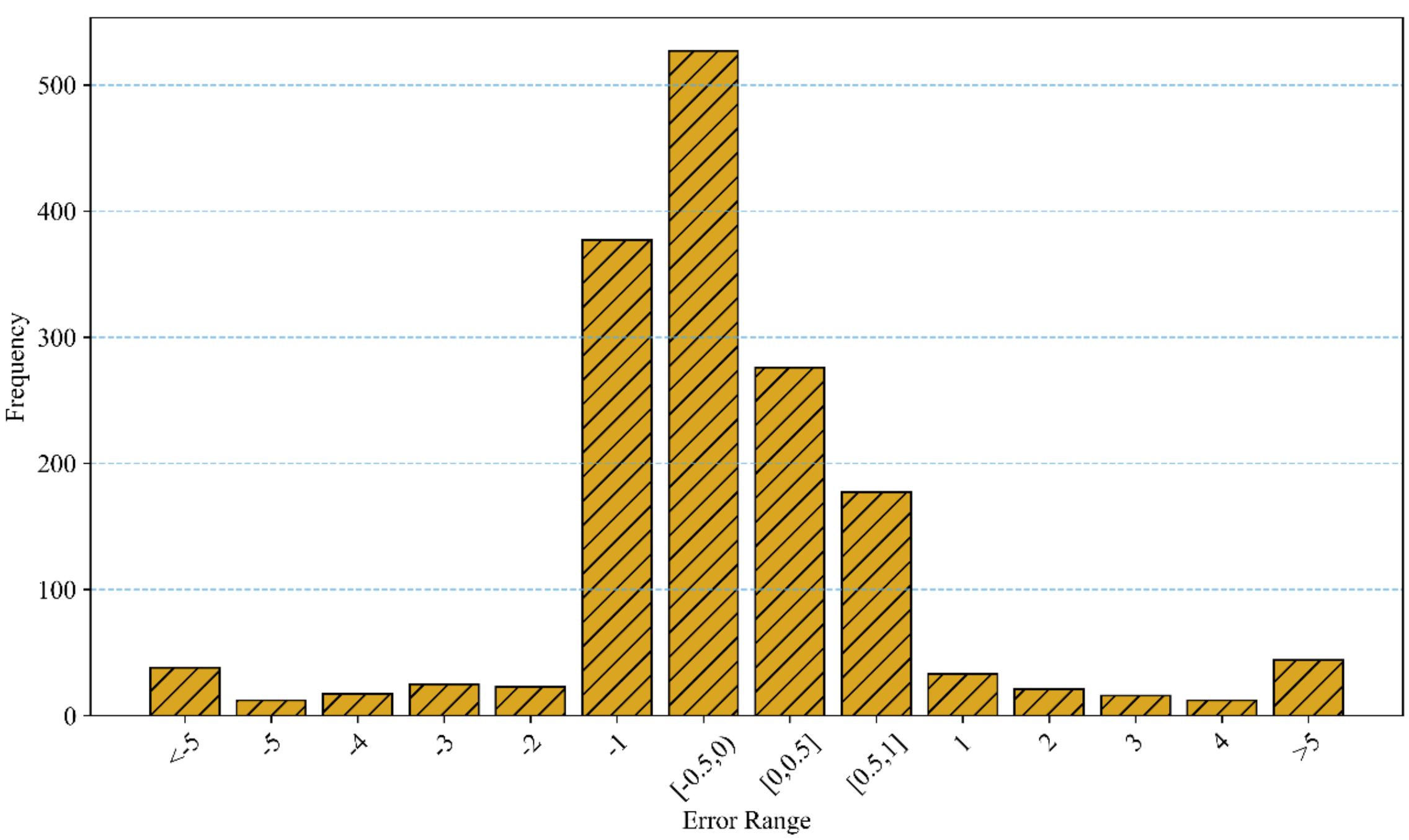

Figure 10. Overall error distribution across the entire dataset (2010–2018)

### 5.3. Feature Importance

After designing a time series architecture with satisfactory predictive performance, the next step involves identifying the features that most significantly impact model predictions. Various methods exist for extracting and evaluating feature importance. In this section, multiple approaches are employed to rank the key features, and their results are compared. One common technique for assessing feature importance is permutation importance, where each feature is sequentially removed from the model and the resulting degradation in model performance is observed. By comparing the changes in evaluation metrics after excluding each feature, it becomes possible to identify which variables exert the greatest influence. For this purpose, several short-term and long-term experiments were conducted. In short-term tests, random months were selected, and the model was evaluated after the removal of individual features. The top 10 features obtained through this method are illustrated in Figure 11.

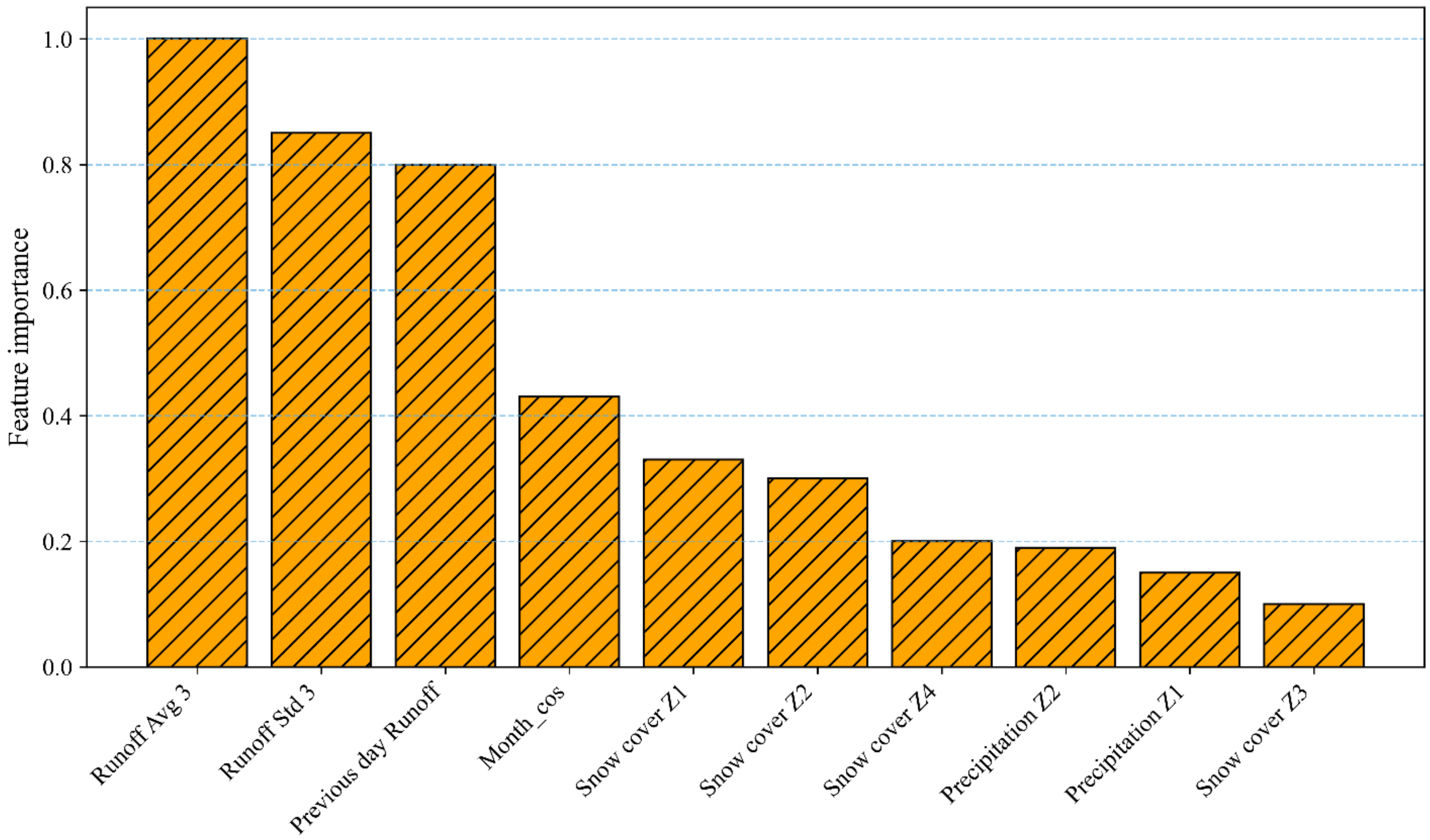

Figure 11. Top 10 influential features identified via permutation importance analysis

Another widely used approach for identifying important features is the Gradient-based Feature Importance technique. This method assesses the sensitivity of the model to input variations in order to determine the relative importance of each feature. The core idea is that the more sensitive the model is to fluctuations in a particular feature, the more significant that feature is in the prediction process. In this approach, a number of test samples are randomly selected, and for each sample, the gradient of the model's output with respect to its input features is computed. These gradients indicate how much a change in each feature affects the model's output. To obtain a quantitative measure of each feature's importance, the mean of the absolute values of these gradients is calculated across the selected samples. Finally, features are ranked based on these averages, with features associated with larger gradient values being identified as more influential. in figure 12 the result of this approach was been reported.

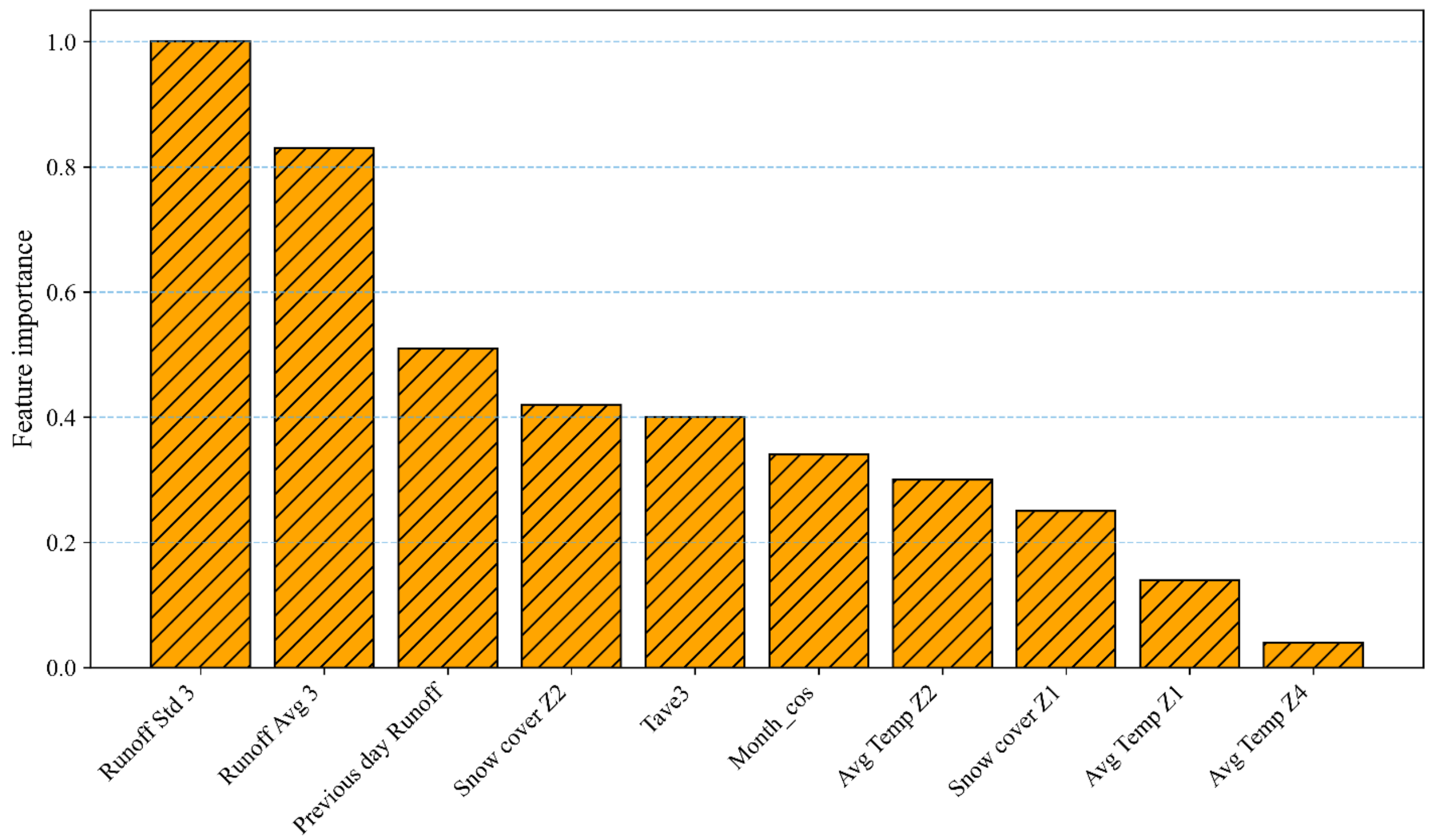

Figure 12. Top 10 Gradient-based feature importance ranking

Another method for identifying influential features involves analyzing the internal structure of LSTM layers and examining the weights of their various gates. In this approach, the importance of each input feature is directly assessed by evaluating its impact on the performance of the LSTM units through the model's trained weights. First, all LSTM layers are extracted from the model. Then, for each layer, the input weight matrices are analyzed. These matrices represent the direct connections between the input features and the different LSTM units. By examining the values within these matrices, it becomes possible to determine the extent to which each input feature contributes to the activation or suppression of memory unit responses. Figure 13 presents the top 10 influential features identified through this method.

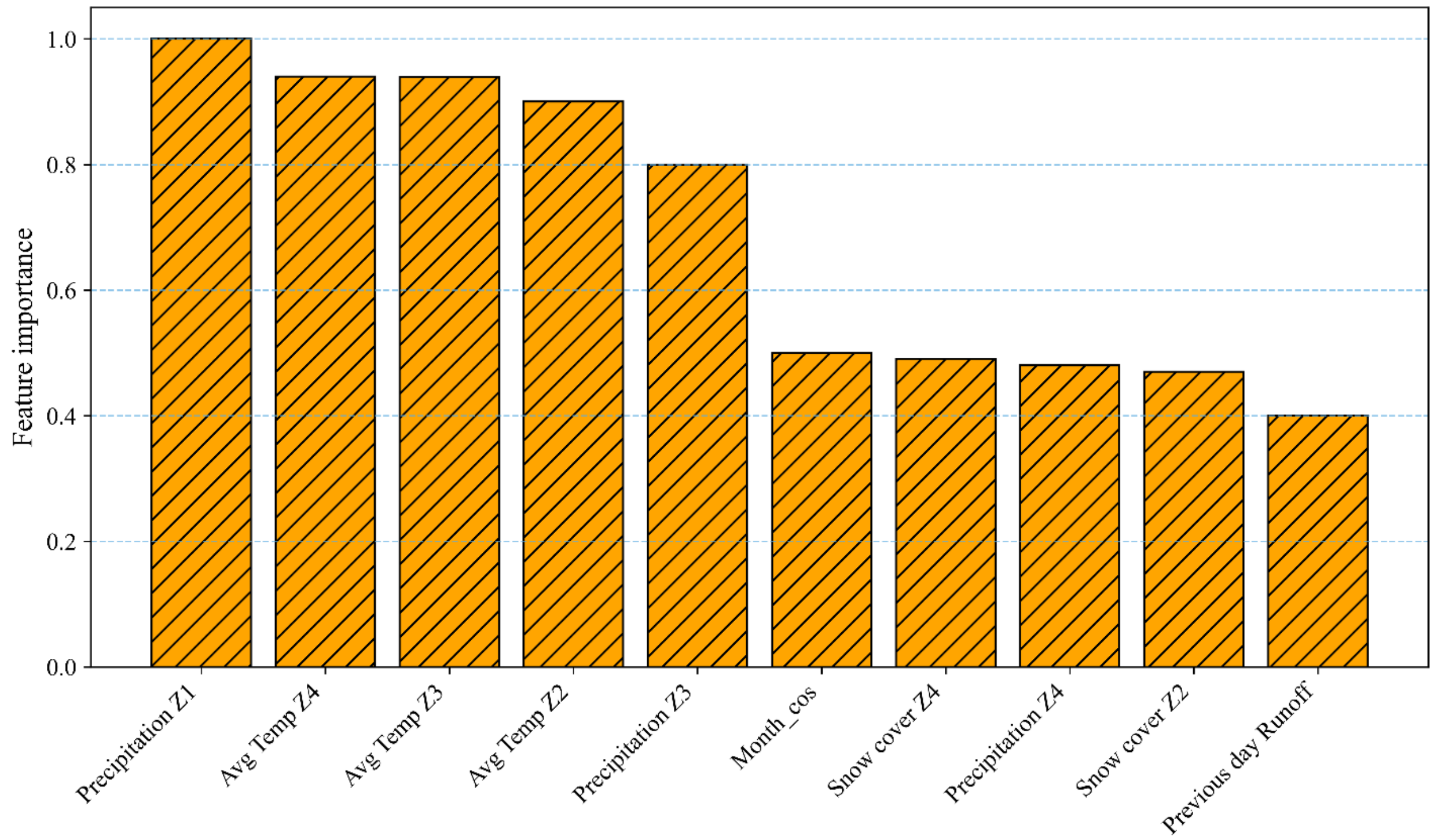

Figure 13. Top 10 Feature importance extracted from LSTM gate weights

It should be noted that in all of these feature importance charts (Figures 9–11), the importance scores were normalized using MinMaxScaler to the range [0, 1] to facilitate easy comparison across different methods.

Finally, a comprehensive comparison of feature importance rankings obtained from the three approaches is provided in Table 3. In this table, the features are ranked separately based on the scores obtained from each method. The table provides an overall perspective on the relative importance of the features within the model and enables a detailed comparison between the outputs of different feature analysis techniques.

Table3. Comparative ranking of feature importance across three methodologies

| **Feature name** | **Methods (Ranking)** | | | |
|---|---|---|---|---|
| | **Permutation (Rank)** | **Gradient-based (Rank)** | **LSTM weights (Rank)** | **Overall (Rank)** |
| Runoff Std 3 | 2 | 1 | 11 | 1 |
| Previous day Runoff | 3 | 3 | 10 | 2 |
| Month_Cos | 4 | 6 | 7 | 3 |
| Avg Temp Z3 | 14 | 5 | 3 | 4 |
| Runoff Avg 3 | 1 | 2 | 15 | 5 |
| Snow cover Z2 | 6 | 4 | 9 | 6 |
| Precipitation Z1 | 8 | 15 | 1 | 7 |
| Snow cover Z1 | 5 | 8 | 12 | 8 |
| Avg Temp Z4 | 15 | 10 | 2 | 9 |
| Avg Temp Z2 | 17 | 7 | 4 | 10 |
| Avg Temp Z1 | 16 | 9 | 5 | 11 |
| Snow cover Z4 | 7 | 17 | 7 | 12 |
| Precipitation Z3 | 12 | 16 | 6 | 13 |
| Precipitation Z4 | 13 | 13 | 8 | 13 |
| Precipitation Z2 | 9 | 14 | 13 | 14 |
| Month_Sin | 11 | 11 | 14 | 15 |
| Snow cover Z3 | 10 | 12 | 16 | 16 |

## 6- Conclusion

In this study, we proposed a hybrid approach to daily runoff prediction by combining LSTM models with different time steps into a single optimized ensemble regressor through the Particle Swarm Optimization (PSO) technique. This allowed the model to learn a broad spectrum of hydrological processes ranging from short-term precipitation effects to slower processes such as snowmelt. Through the tuning performed by PSO, the ensemble model exhibited excellent predictive capability with $R^2$ of up to 91.42%. By employing several temporal resolutions, the model broke the conventional constraints of one-step LSTMs in providing a more effective and adaptive way of capturing temporal dependencies in runoff data. Our feature importance study via permutation techniques, gradient-based attribution, and LSTM gate weights revealed the most important inputs as short-term runoff variability, past discharge quantities, and seasonality. Further, spatial inputs such as temperature and snow cover by elevation bands highlighted microclimate control on runoff dynamics.

**Funding:** The authors declare that no funds, grants, or other support were received during the preparation of this manuscript.

**Author Contributions Statement:**

H.S.: Supervision, Review & Editing.

A.E.: Coding, data analysis, Writing – Original Draft, Editing.

B.H.: Writing, investigation

**Competing Interests:**

The authors declare that they have no conflict of interest.